%% file: main.tex
\documentclass[10pt,twocolumn,letterpaper]{article}
\usepackage{newtxtext}
\usepackage[pagenumbers]{cvpr} 
\usepackage{multirow} 
\usepackage{siunitx}
\usepackage{pifont}
\newcommand{\cmark}{\textcolor{green}{\ding{51}}}
\newcommand{\xmark}{\textcolor{red}{\ding{55}}} 
\usepackage{xcolor}
\usepackage{xspace}
\usepackage{indentfirst}
\usepackage{comment}
\usepackage{tikz}
\usepackage[accsupp]{axessibility}
\usepackage{textcomp}
\usepackage{gensymb}
\definecolor{cvprblue}{rgb}{0.21,0.49,0.74}
\usepackage[pagebackref,breaklinks,colorlinks,allcolors=cvprblue]{hyperref}

\newcommand{\methodName}{\textsc{Trophies}\xspace}

\title{\methodName: \underline{T}emporal \underline{R}econstruction \underline{o}f \underline{P}laces, \underline{H}umans, and Cameras \\ from Mult\underline{i}-vi\underline{e}w Video\underline{s}}

\author{Jinpeng Liu \qquad Yukang Xu \qquad Yutong Li \qquad Xingyu Liu\\National University of Singapore}

\newcommand{\titleemoji}{%
  \raisebox{-0.1em}{\includegraphics[height=1em]{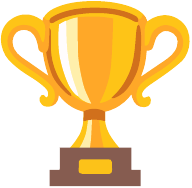}}%
  \hspace{0.2em}%
}

\makeatletter
\patchcmd{\@maketitle}
  {{\Large \bf \@title \par}}
  {{\Large \bf \titleemoji \@title \par}}
  {}{}
\patchcmd{\@maketitle}
  {{\large \bf \@title \par}}
  {{\large \bf \titleemoji \@title \par}}
  {}{}
\makeatother

\begin{document}
\twocolumn[{
    \renewcommand\twocolumn[1][]{#1}
    \maketitle
    \begin{center}
        \vspace{-20pt}
        \includegraphics[width=1.0\linewidth]{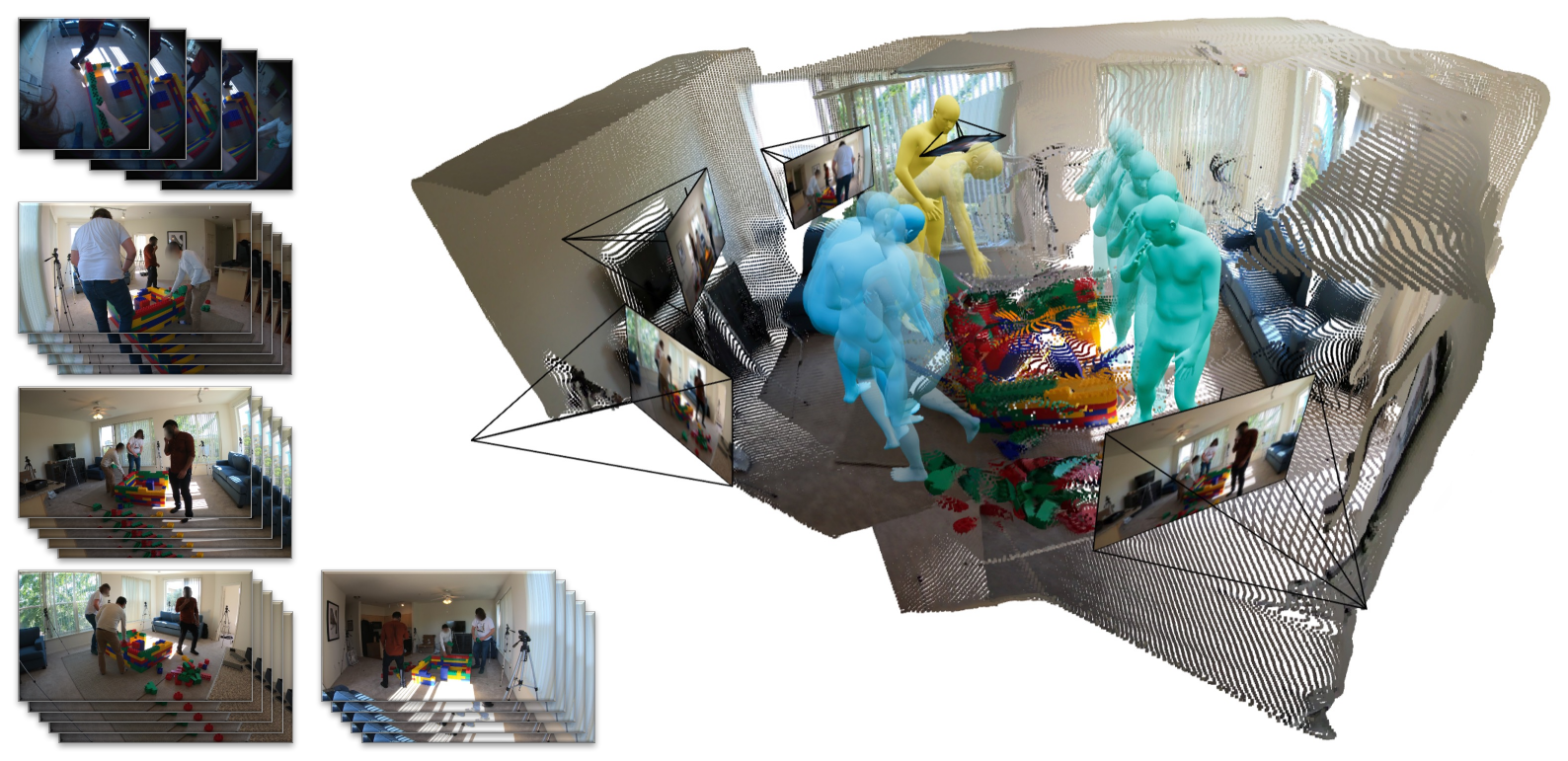}
        \vspace{-20pt}
        \captionof{figure}{\textbf{Overview of \methodName.} 
        Given temporally synchronized video streams, \methodName jointly reconstructs dynamic humans, static scene geometry, and camera trajectories within a globally consistent 4D space. Our method couples a human branch and a scene branch through a global alignment and optimization stage that enforces scale, contact, and gravity consistency. This unified reconstruction produces temporally stable and spatially coherent human–scene representations, where human motions, scenes, and camera viewpoints are all aligned in a shared world coordinate frame. }
        \label{fig:teaser}
    \end{center}
}]

\input{sec/0_abstract}
\input{sec/1_intro}
\input{sec/2_releated}
\input{sec/3_1_task}

\input{sec/3_method}
\input{sec/4_experiments}
\input{sec/5_con}
\input{sec/6_ack}

{
    \small
    \bibliographystyle{ieeenat_fullname}
    \bibliography{main}
}

\end{document}

%% file: sec/0_abstract.tex
\begin{abstract}
Reconstructing humans and their surrounding environments in a globally consistent 4D space is essential for comprehensive perception. However, prior works typically assume single-view inputs or decouple humans, scenes, and cameras, making them unable to recover coherent geometry, stable motion, and physically aligned trajectories. These limitations motivate us to introduce a new task: unified human–scene–camera reconstruction from multi-view videos, which aims to jointly estimate dynamic humans, static scenes, and camera poses in one global coordinate frame. We propose \methodName\  -- \textit{Temporal Reconstruction of Places, Humans, and Cameras from Multi-view Videos} -- a unified framework tailored for this task. \methodName features a Human Branch that models human through temporal and spatial reasoning, and a Scene Branch that reconstructs static geometry with human-aware attention. A global alignment and optimization module couples both branches by enforcing scale consistency, contact priors, and cross-view temporal coherence. Experiments on EgoHuman and EgoExo4D demonstrate that \methodName achieves globally aligned, physically plausible 4D reconstructions and consistently outperforms existing paradigms in both global fidelity and human–scene consistency.
\end{abstract}

%% file: sec/1_intro.tex
\vspace{-10pt}
\section{Introduction}
Understanding how humans move and interact within complex 3D environments is a long-standing goal in computer vision and embodied AI~\cite{deepmotion, realhumans, scalingHCI, unifiedHCI, Human-X-Interaction}. Significant advances have been made in human motion modeling~\cite{flag3d, flag3d++, scalingHCI, prox, yukangstaf, yukangreconstructing, yukangsdasd} and 3D scene reconstruction~\cite{monst3r, mast3r, dust3r, vggt, easi3r}, yet these developments largely reconstruct \emph{disconnected worlds}. Human estimators predict articulated motion in local coordinate systems that drift over time, while scene reconstruction pipelines recover geometry only up to scale and often ignore dynamic subjects. This separation prevents a physically consistent understanding of human–scene interaction and limits downstream applications in AR/VR, robotics, and embodied intelligence.

Existing methods address only isolated aspects of the full challenge. Single-view human motion estimators~\cite{hmr2, tram, gvhmr, wham} excel at capturing articulated pose but lack awareness of the surrounding environment and camera motion. Geometry-based methods~\cite{dust3r, mast3r, droid, vggt-slam} reconstruct scenes and camera trajectories but assume static environments, and typically fail in the presence of human motion. Efforts such as JOSH~\cite{josh} and Human3R~\cite{human3r} jointly infer humans and scenes but remain limited to single-view inputs and rely heavily on priors, resulting in inconsistent scales and physically implausible interactions. Recent multi-view approaches like HSfM~\cite{hsfm} only operate per frame and yield inconsistent dynamics over time. These limitations highlight a core challenge: there is still no unified framework that reconstructs humans, scenes, and cameras together in a globally consistent 4D world. 
Such a task requires integrating signals across geometry, motion, and physical constraints, while aligning multiple coordinate systems into a unified reference frame.

In this paper, we adopt a multi-view formulation of this task because multi-view observations provide stronger geometric cues and reduce scale ambiguity. We emphasize that our formulation is view-agnostic: the architectural components of our approach do not rely on multi-view supervision. As a result, our framework seamlessly adapts to single-view videos. When only one view is available, it reduces to a temporally coherent and scale-stable human–scene–camera reconstruction system, maintaining its flexibility while preserving physical consistency. This generality allows our method to handle multi-camera systems, sparse camera arrays, and handheld monocular inputs.

To tackle the challenge of unified reconstruction, we introduce a framework that jointly reasons about scene geometry, human motion, and camera motion within a single optimization loop (\cref{fig:pipeline}). Our scene branch is designed as a plug-and-play module compatible with geometry estimators such as DUSt3R~\cite{dust3r}, MonST3R~\cite{monst3r} and CUT3R~\cite{cut3r}. It incorporates a human-aware attention mechanism that suppresses dynamic regions by modulating attention weights rather than relying on explicit segmentation~\cite{easi3r}. For CUT3R, we further introduce a multi-memory design that disentangles static and dynamic elements across time and views, enabling robust recovery of scene geometry and camera trajectories even in the presence of substantial human motion. In parallel, our human branch leverages a multi-view transformer with symmetric and anchor-referenced attention to achieve cross-view consistent reconstruction of body pose and motion. Finally, the alignment and optimization module links both branches through Sim(3) alignment, scale calibration, and contact-aware optimization, producing physically grounded reconstructions in a unified world coordinate system.

In summary, these components establish a unified formulation for reconstructing humans, scenes, and cameras under a physical world temporally. By resolving long-standing inconsistencies in scale, dynamics, and coordinate alignment, our approach moves beyond fragmented pipelines toward holistic 4D understanding.

\input{tab/different_feature}

%% file: tab/different_feature.tex
\begin{table}[t]
\centering
\caption{\textbf{Comparison of methods across different features.}
}
\vspace{-5pt}
\scriptsize
\resizebox{0.47\textwidth}{!}{
\begin{tabular}{l|cccccc}
\hline
 & Places & Human Pose & Camera & Temporal &  Multi-view & Gravity-aware \\ \hline 
HMR2~\cite{hmr2, hmr} & \xmark & \cmark   &   \xmark  &  \cmark  & \xmark & \xmark \\
TRAM~\cite{tram} &  \xmark  &   \cmark    &   \xmark    &   \cmark &  \xmark  & \xmark \\
GVHMR~\cite{gvhmr} &  \xmark  &   \cmark    &   \xmark    &   \cmark &  -  & \cmark \\
DUSt3R~\cite{dust3r} & \cmark  &   \xmark    &  \cmark   &  - &  \cmark  & \xmark \\
CUT3R~\cite{cut3r} &  \cmark  &  \xmark   &  \cmark  &  - &  \cmark  &  \xmark \\
JOSH~\cite{josh} & \cmark  & \cmark  & \cmark  & \cmark  &  \xmark  & \xmark    \\
Human3R~\cite{human3r} &    \cmark  &    \cmark   &   \cmark   & \cmark  &         \xmark  &   \xmark    \\
HSFM~\cite{hsfm} &    \cmark  &    \cmark   &   \cmark  & \xmark  &  \cmark   &   \xmark   \\ \hline 
Ours  &   \cmark &   \cmark &   \cmark  &  \cmark  &   \cmark    &     \cmark      \\ \hline 
\end{tabular}
}
\vspace{-15pt}
\label{tab:diff-feature}
\end{table}

%% file: sec/2_releated.tex
\section{Related Work}
\textbf{Human Reconstruction and Motion Estimation.}
Early single-view methods such as HMR~\cite{hmr}, SPIN~\cite{spin}, and PARE~\cite{pare} estimate SMPL parameters from individual frames but suffer from scale ambiguity and temporal jitter. Temporal models like VIBE~\cite{vibe}, TCMR~\cite{tcmr}, and TMR~\cite{tmr} improve motion smoothness via recurrent or transformer architectures, yet lack spatial grounding. Some works~\cite{josh, rich, human3r} predict contact labels for contact aware human reconstruction, yet remain monocular. Recent works, including TRAM~\cite{tram}, WHAM~\cite{wham}, SLAHMR~\cite{slahmr}, GVHMR~\cite{gvhmr}, and VideoMimic~\cite{VideoMimic} infer world-aligned human trajectories from monocular inputs by jointly reasoning about motion and implicit camera pose, but still cannot resolve cross-view ambiguities or guarantee multi-camera consistency. In contrast, our Human Branch is explicitly designed for true multi-view inputs, producing temporally stable and geometrically consistent SMPL~\cite{SMPL, SMPL-X, MANO}.

\textbf{Scene Reconstruction and Multi-View Geometry.}
Parallel to human motion estimation, reconstructing static scenes from multiple views has been a central problem in 3D vision. Traditional Structure-from-Motion (SfM)and Multi-View Stereo (MVS) pipelines~\cite{sfm, mvs} recover accurate geometry through feature matching and triangulation, but they typically fail under dynamic or low-texture conditions. Neural implicit methods such as NeRF~\cite{nerf} and NeuS~\cite{neus} improve robustness and surface quality by learning continuous volumetric fields, yet they require dense view sampling and struggle with moving humans. Recent transformer-based approaches, such as DUSt3R~\cite{dust3r}, MASt3R~\cite{mast3r}, and CUT3R~\cite{cut3r}, introduce dense feature matching to jointly estimate camera poses and scene depth. These methods provide strong geometric priors for multi-view reconstruction but assume static environments and thus deteriorate in the presence of dynamic human motion. Our Scene Branch introduces a human-aware attention mechanism that modulates cross-view interactions. This design enables the model to preserve static background consistency while filtering out dynamic human regions.

\textbf{Joint Human–Scene Reconstruction.}
Jointly recovering humans and their environments is crucial for understanding dynamic scenes. Early works such as PROX~\cite{prox} and BEHAVE~\cite{behave} leverage scene geometry via contact or collision priors but depend on pre-scanned static environments without reconstructing full scenes. Later approaches, including PHSP~\cite{populating}, HPS~\cite{HPS}, and InterCap~\cite{2022intercap, 2024intercap}, jointly optimize human models with reconstructed environments, yet typically rely on single-frame inputs or precomputed meshes, limiting temporal consistency. More recent methods like HSC4D~\cite{hsc4d}, H4D~\cite{h4d}, JOSH~\cite{josh}, and HSFM~\cite{hsfm} achieve more coherent 4D human–scene understanding by integrating motion cues, physical constraints, but still treat human and scene components in loosely coupled stages, resulting in scale drift and inconsistent global alignment. In contrast, our method introduces a globally aligned human–scene optimization framework that tightly couples both reconstruction branches through Sim(3) alignment, contact-aware constraints, and bundle-adjustment refinement, enabling globally consistent 4D reconstruction.

%% file: sec/3_1_task.tex
\section{Preliminaries and Formulation}
\begin{figure}
    \centering
    \includegraphics[width=1.0\linewidth]{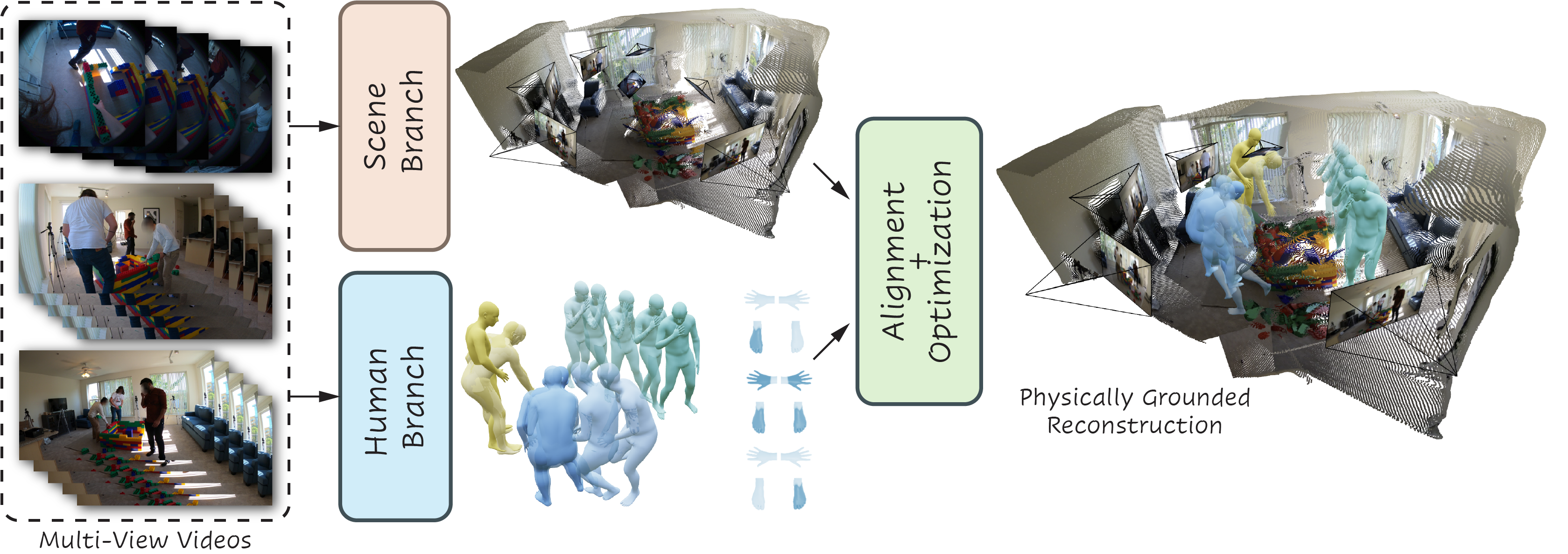}
    \vspace{-20pt}
    \caption{\textbf{Pipeline Overview.} Our framework consists of three components: a Scene Branch that reconstructs the static environment with human-aware attention (implemented as a plug-and-play module applicable to DUSt3R, MonST3R, and CUT3R backbones); a Human Branch that estimates temporally coherent body parameters from multi-view videos via symmetric and anchor-referenced attention; and a global Align and Optimization stage that unifies humans, scenes, and cameras under consistent geometry and contact-aware constraints, producing a physically grounded reconstruction in a shared world coordinate system.}
    \vspace{-10pt}
    \label{fig:pipeline}
\end{figure}

\textbf{Setup.} 
Our model takes a set of uncalibrated multi-view videos as input. We denote the $N$ synchronized video streams as 
$\{V^n\}_{n=1}^{N}$, where each video 
$V^n = \{I_1^n, I_2^n, \ldots, I_T^n\}$ contains $T$ temporally ordered frames captured by camera $n$. Given this input, our method jointly reconstructs humans, scenes, and cameras in a metric 3D world over time. The output consists of temporally coherent human reconstructions ${H_t^h}$, static scene representations ${\mathcal{S}_t^n}$, and camera parameters ${R_t^n, t_t^n, \alpha_t}$, together forming a consistent 4D representation that serves as the foundation for our joint human–scene–camera optimization framework.

\textbf{Human.}
We represent each human in the scene using the SMPL~\cite{SMPL, SMPL-X} body model.
For a human $h$ at time $t$, the body parameters are defined as
\begin{equation}
    H_t^h = \{\phi_t^h, \theta_t^h, \beta^h, \gamma_t^h\},
\end{equation}
where $\phi_t^h \in SO(3)$ denotes the global orientation, 
$\theta_t^h \in SO(3)^J$ the articulated joint rotations, 
$\beta^h \in \mathbb{R}^B$ the shape coefficients, and 
$\gamma_t^h \in \mathbb{R}^3$ the global translation.
Unlike single-frame or single-view methods, our multi-view video setup maintains a shared shape $\beta^h$ across all frames and enforces temporal smoothness on $\{\phi_t^h, \theta_t^h, \gamma_t^h\}$, yielding a temporally coherent 4D body trajectory.

\textbf{Scene.} 
We represent the static environment as a set of dense 3D point maps reconstructed from multi-view geometry~\cite{dust3r, monst3r, cut3r}.
For view $n$ and frame $t$, we define
$\mathcal{S}_t^n \in \mathbb{R}^{H \times W \times 3}$,
where each pixel corresponds to a 3D point in the world coordinate frame.
Given the intrinsic matrix $K^n$ and per-frame extrinsic matrix $(R_t^n, t_t^n)$, a pixel $(i, j)$ with depth value $D_{i, j}^{n, t}$ is unprojected to the world via
\begin{equation}
    \mathcal{S}_{i, j}^{n,t} = 
    \alpha_t (R_t^{n\top} [(K^n)^{-1} D_{i,j}^{n,t} [i, j, 1]^\top - t_t^n]),
\end{equation}
$\alpha_t$ is a scaling factor that aligns points in the metric space.

\textbf{Cameras.}
Each camera $n$ is modeled by intrinsic parameters $K^n$ and extrinsic matrix $(R_t^n, t_t^n)$.
We introduce a similarity scale $\alpha_t$ that normalizes inter-camera distances across time.
A 3D point $x_{3D}$ is projected to camera $n$ at frame $t$ as
\begin{equation}
    x_{2D}^{n,t} = K^n (R_t^n x_{3D} + \alpha_t t_t^n).
\end{equation}

%% file: sec/3_method.tex
\section{Method}
We propose a framework that decomposes the reconstruction into three collaborative components, as illustrated in~\cref{fig:pipeline}. Scene Branch focuses on recovering the static background geometry and camera poses under a human-aware attention mechanism. Human Branch estimates body parameters from synchronized multi-view inputs through a symmetric and anchor-referenced attention design. The final module bridges both branches via alignment and contact-aware constraints, unifying humans, scenes, and cameras within a coherent world coordinate frame. 

\begin{figure}
    \centering
    \includegraphics[width=1.0\linewidth]{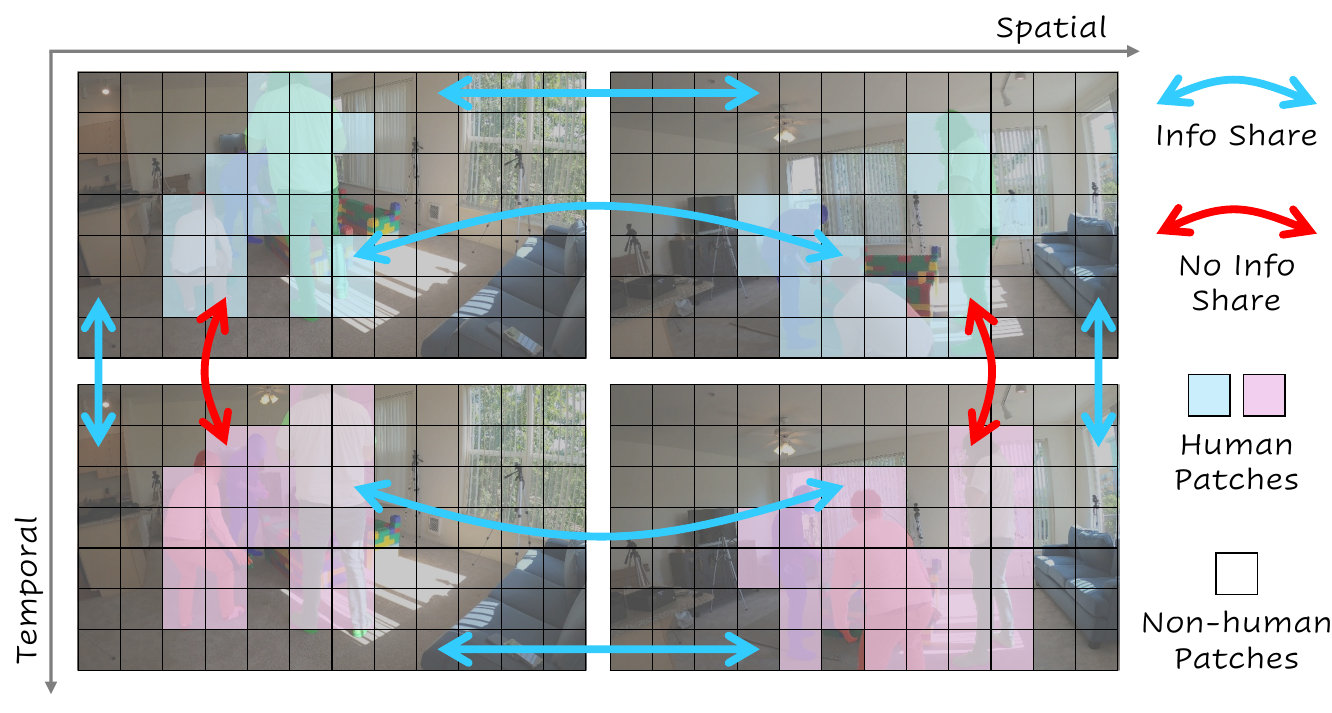}
    \vspace{-20pt}
    \caption{
    \textbf{Human-aware attention.} Each frame is divided into human (colored) and non-human (transparent) patches. For views at the same time, all patches share information to enforce multi-view consistency. Across different time steps, only non-human patches exchange information, while human patches are masked in the attention layer to avoid motion-induced inconsistency. 
    }
    \label{fig:scene-branch}
    \vspace{-10pt}
\end{figure}

\begin{figure*}
    \centering
    \includegraphics[width=0.9\linewidth]{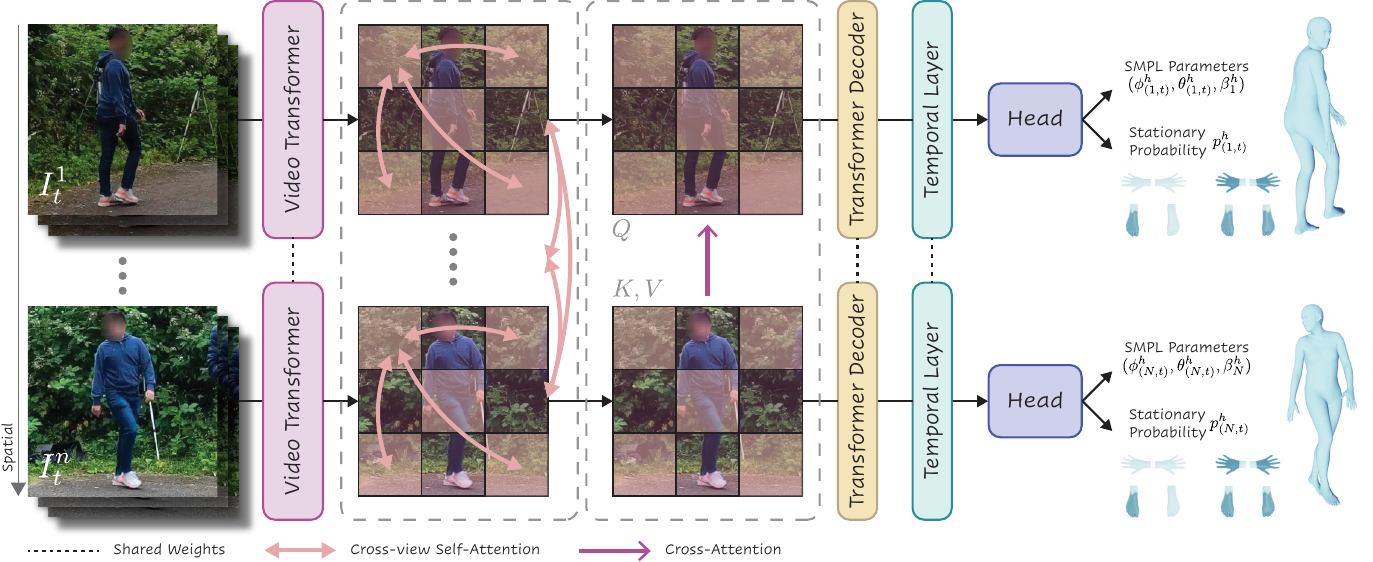}
    \caption{Our method takes synchronized multi-view video frames as input and processes them with shared Human Video Transformers. A two-stage cross-view module first performs symmetric attention among all views to exchange global geometric information, and then applies an anchor-referenced fusion that aggregates features into the anchor view. The fused representation is decoded temporally and passed to dual heads predicting SMPL parameters $(\phi_{(n,t)}^h, \theta_{(n,t)}^h, \beta_{n}^h )$ and stationary contact probabilities ${p^h_{(n,t)}}$. During inference, the anchor view’s 3D joints are used as the final output, embedding multi-view consistency and contact awareness.}
    \label{fig:human-branch}
    \vspace{-10pt}
\end{figure*}

\subsection{Scene Branch}\label{sec:scene}
To recover the surrounding static environment from human videos, we design a plug-and-play module that can be seamlessly integrated into recent dense multi-view reconstruction frameworks such as DUSt3R~\cite{dust3r}, MonST3R~\cite{monst3r} and CUT3R~\cite{cut3r}. This module introduces human-aware attention, implemented through either mask-based attention reweighting~\cite{dust3r, monst3r, easi3r} or multi-memory reasoning~\cite{cut3r}, depending on the backbone architecture as illustrated in~\cref{fig:scene-branch}. These mechanisms collectively enable stable scene recovery from dynamic multi-view videos by suppressing transient human motion while preserving consistent static geometry. For DUSt3R and MonST3R, given images ${I_{t_1}^a}$ and ${I_{t_1}^b}$ from different perspectives ($a$ and $b$) at the same time $t_1$, the underlying backbone $\mathcal{B}$ predicts dense 3D point maps $\mathbf{P}_{t_1}^{a\leftarrow b}$ and camera parameters $\mathbf{C}_{t_1}^a$:
\begin{equation}
\begin{split}
    (\mathbf{P}_{t_1}^{a\leftarrow b}, \mathbf{C}_{t_1}^a) = \mathcal{B}(I_{t_1}^{a}, I_{t_1}^{b}),
\end{split}
\end{equation}
where all the tokens from view $b$ contribute to view $a$.
Given images ${I_{t_1}^a}$ and ${I_{t_2}^b}$ from different viewpoints ($a$ and $b$) at different times ($t_1$ and $t_2$), the underlying backbone $\mathcal{B}$ predicts dense 3D point maps: 
\begin{equation}
(\mathbf{P}_{t_1}^{a\leftarrow b}, \mathbf{C}_{t_1}^a) = \mathcal{B}(I_{t_1}^{a}, I_{t_2}^{b}, M_{human}^{a\leftarrow b}),
\end{equation}
where $M_{human}^{a\leftarrow b} = (1-M_{human}^a)\otimes (M^b_{human})^T$, $\otimes$ denotes the outer product and $M_{\text{human}}$ is a binary mask indicating patches that contain human. This results in tokens from human regions from viewpoint $b$ that do not contribute to static regions from viewpoint $a$. We implement it through attention reweighting to suppress human regions:
\begin{equation}
softmax^{a\leftarrow b}(\mathbf{\hat{A}}) =\left\{\begin{matrix}
 0 & \text{if}~M_{\text{human}}^{a\leftarrow b} \\
  \text{softmax}(A^{a\leftarrow  b}) & \text{otherwise}
\end{matrix}\right.
\end{equation}
where $\mathbf{A}^{a\leftarrow b}$ denotes the attention matrix.

Due to CUT3R's memory mechanism, the idea of human-aware attention cannot be achieved through above methods. We adopt a multi-memory disentanglement strategy: one memory bank aggregates multi-view human and scene features at a single moment for spatial consistency, while another stores static scene features across time to maintain temporal stability. 

\subsection{Human Branch}\label{sec:human}
Given synchronized video streams ${I^n_t}$ from different camera views $n$, our human branch predicts per-frame SMPL parameters and stationary probability maps indicating human–scene contact as shown in~\cref{fig:human-branch}. 

For each input view $n$, we employ a Human Video Transformer backbone~\cite{tram} that extracts spatio-temporal tokens from short frame sequences. The backbone operates on individual views but shares weights across all cameras to maintain view-consistent feature distributions.

To exploit multi-view consistency, we introduce a two-step cross-view interaction module. Unlike previous single-view methods~\cite{slahmr, tram, wham, gvhmr, yukangstaf}) which only model temporal dependencies, our cross-view attention~\cite{lgm} explicitly aligns spatial features across cameras by allowing each patch token to attend to its counterparts in other views. This enhances geometric awareness and implicitly enforces view-consistent body reconstruction, without relying on explicit 3D supervision. In the first step, all views interact symmetrically, where each view’s tokens attend to those of every other view. This allows the network to capture geometric correspondences and complementary visibility cues across cameras. This operation yields multi-view contextualized features $F'_n$ for all views.

Following this symmetric exchange, we perform an anchor-referenced cross-view fusion to consolidate information into a single representative view.
A designated anchor view (typically the frontal or most stable camera) provides the queries (Q), while the remaining reference views provide the keys (K) and values (V):
\begin{equation}
    F^{''}_{\text{anchor}} = \text{softmax}(\frac{Q_{\text{anchor}}K^T_{\text{ref}}}{\sqrt{d}})V_{\text{ref}}.
\end{equation}

This fusion enriches the anchor view's representation with complementary geometry and appearance cues from the reference views while preserving its spatial structure. During training, the anchor view is randomly sampled from the available views for generalization. For inference, we select the view with the highest human detection confidence. Only the anchor view’s 3D joints are used as the model output for inference, as they already integrate the fused multi-view information through this two-step attention process.
Together, the symmetric global interaction and anchor-referenced refinement provide a compact yet powerful mechanism for multi-view feature integration, balancing geometric completeness and computational efficiency. 
The fused anchor features are then decoded by a Transformer Decoder and a Temporal Layer that captures short-term motion dynamics. 
The hierarchical modeling -- first across views (symmetric + anchor fusion) and then across time -- ensures stable and smooth pose reconstruction over sequences. A dual-head prediction module outputs: (1) SMPL Parameters $(\phi_{(n,t)}^h, \theta_{(n,t)}^h, \beta_{n}^h)$, representing orientation, pose, shape; (2) Stationary Probability Maps $p_{(n,t)}^h$ indicating per-region contact likelihood with the surrounding scene. These contact maps introduce physical constraints and later guide the optimization in ~\cref{sec:align}.

\subsection{Alignment and Optimization}\label{sec:align}
The final stage of our framework performs a global alignment and optimization that unifies human, scene, and camera representations within a consistent 3D space. 
We design this process that successively enforces metric consistency~\cite{zoedepth, droid}, scene refinement~\cite{dust3r, cut3r, monst3r}, physical grounding~\cite{rot6d, ransac}, and multi-view coherence~\cite{lgm}.

\subsubsection{Alignment} 
Given a set of camera poses $\{[\mathbf{R}_i, \mathbf{T}_i]\}$ and image depth maps $\{D_i\}$, our goal is to estimate a similarity transformation $S_i = [s_i, \mathbf{R}_i', \mathbf{T}_i'] \in \mathrm{Sim(3)}$ that brings each external estimate into the coordinate frame of the Scene Branch. For each view $i$, we solve:
\begin{equation}
S_i = 
\mathop{\arg\min}_{S \in \mathrm{Sim(3)}} 
\sum_{\mathbf{x} \in \Omega_i}
\left\|
S \cdot K^{-1}[\mathbf{x}, D_i(\mathbf{x})] 
- \mathbf{P}_i(\mathbf{x})
\right\|_2^2,
\label{eq:sim3}
\end{equation}
where $K^{-1}[\mathbf{x}, D_i(\mathbf{x})]$ denotes the back-projection of pixel $\mathbf{x}$ using its depth, and $\mathbf{P}_i(\mathbf{x})$ represents the corresponding 3D points reconstructed by the Scene Branch.

For dynamic-camera sequences, we first estimate the relative camera motion using DROID-SLAM~\cite{droid}, which provides accurate frame-to-frame poses up to an unknown global scale. To recover the absolute metric scale, we employ ZoeDepth~\cite{zoedepth} to predict depth maps $D_i$ for each frame. The resulting pair of relative poses and absolute depths allows us to optimize~\cref{eq:sim3}, aligning the DROID-SLAM trajectory with the Scene Branch reconstruction and resolving the inherent scale ambiguity of monocular SLAM.

For static-camera setups, where camera poses are fixed and relative motion is absent, we directly use ZoeDepth to estimate per-view metric depths. The same Sim(3) alignment in~\cref{eq:sim3} is then applied to register all static views with the Scene Branch, ensuring a consistent global scale and coordinate origin across all views.

This unified alignment procedure handles both dynamic and static capture scenarios under a single optimization objective, producing a metrically consistent and globally aligned reference frame for subsequent joint optimization of human, scene, and camera parameters.
\begin{figure*}
    \centering
    \includegraphics[width=1.0\linewidth]{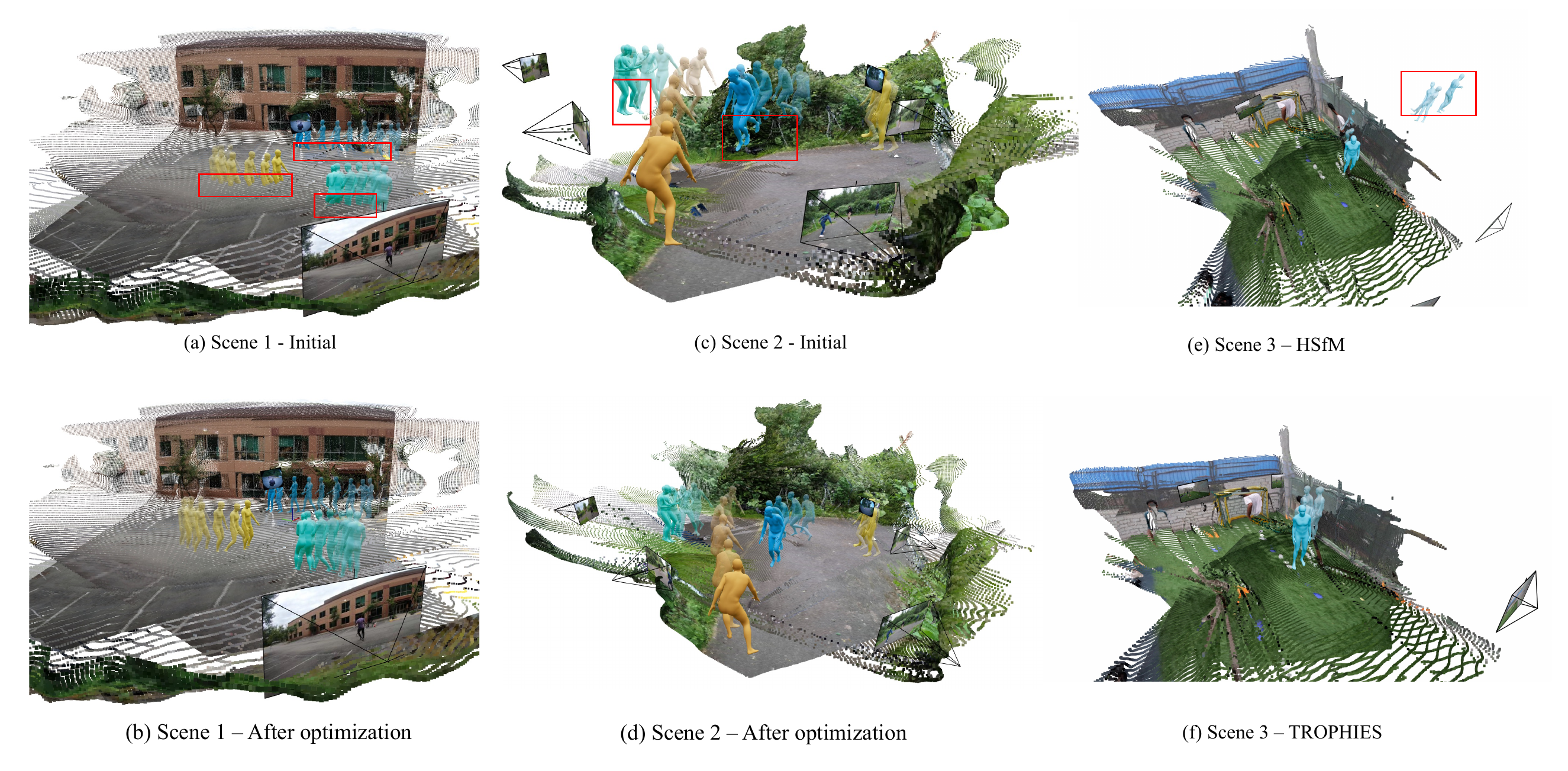}
    \vspace{-20pt}
    \caption{\textbf{Qualitative results.} 
    Comparison of multi-view reconstructions before and after global optimization (Scenes~1–2) and against prior work (Scene~3). For Scenes~1 and~2, the initial results (a,c) exhibit misalignment between humans and scenes, leading to interpenetration, floating feet, and incorrect grounding (red boxes). Our global optimization (b,d) lead to physically coherent and well-grounded reconstructions. 
    Scene~3 compares HSfM (e) with our \methodName framework (f). Since HSfM performs frame-wise independent optimization, its per-frame scale drifts accumulate over time, causing noticeable spatial inconsistencies when the results are aggregated. In contrast, our \methodName maintains a globally consistent scale across the entire sequence, producing stable and coherent human motion.}
    \label{fig:opt}
    \vspace{-10pt}
\end{figure*}

\subsubsection{Optimization}
After achieving global alignment across camera sources, we jointly refine the geometry and motion parameters through unified optimization that includes scene-human bundle adjustment and contact-aware optimization.

\textbf{Bundle Adjustment.}
To reconcile dense scene geometry with articulated human motion, we perform a joint optimization that refines both the camera parameters and the human parameters. We extend the bundle adjustment formulation~\cite{dust3r, hsfm} to minimize re-projection error and consistency losses over all frames and views:
\begin{equation}
\mathcal{L}_{\text{BA}} = \frac{1}{NT}\sum_{n,t}\mathcal{L}_{\text{Scene}} +  \frac{1}{NTH}\sum_{n,t,h}\mathcal{L}_{\text{Human}}
\label{eq:joint_ba}
\end{equation}
\begin{equation}
\mathcal{L}_{\text{Scene}} =
\| 
K^{(n,t)} [\mathbf{R}^{(n,t)}, \mathbf{T}^{(n,t)}]
\mathbf{P}_{\text{scene}} - \mathbf{x}^{(n,t)}_{\text{scene}}\|_2^2
\end{equation}
\begin{equation}
\begin{split}
\mathcal{L}_{\text{Human}} = 
\| 
K^{(n,t)} [\mathbf{R}^{(n,t)}, \mathbf{T}^{(n,t)}]
\text{SMPL}(H^{(n,t,h)}) - \mathbf{J}_{2D}^{(n,t,h)}\|_2^2
\end{split}
\end{equation}
where 
$\mathbf{J}_{2D}^{n,t,h}$ 
are 2D human joints from detection model~\cite{xu2022vitpose, xu2022vitpose+}. The optimization jointly updates camera extrinsics, scene geometry, and human poses to minimize cross-view reprojection errors, thereby achieving globally consistent reconstruction under the unified coordinate system established in~\cref{eq:sim3}.

\textbf{Contact-Aware Optimization.}
To further enforce realistic physical grounding, we introduce a contact-aware term that aligns the reconstructed human motion with the underlying scene geometry. Based on the aligned gravity direction~\cite{SPEC} and the estimated ground plane~\cite{ransac}, we define a set of potential contact vertices $\mathcal{C}$ on the feet and hands of the SMPL mesh. We encourage these vertices to remain close to the scene surface while penalizing interpenetration:
\begin{equation}
\begin{split}
\mathcal{L}_{\text{contact}} =
\sum_{\mathbf{v} \in \mathcal{C}} 
\big(
w_c \cdot \text{dist}(\mathbf{v}, \mathcal{S}_{\text{surface}})^2
+ \\ w_p \cdot \max(0, -\text{n}_{\mathcal{S}}^\top (\mathbf{v} - \mathbf{p}_{\mathcal{S}}))^2
\big),
\end{split}
\end{equation}
where $\text{dist}(\mathbf{v}, \mathcal{S}_{\text{surface}})$ denotes the shortest distance from vertex $\mathbf{v}$ to the scene surface $\mathcal{S}_{\text{surface}}$, $\text{n}_{\mathcal{S}}$ is the surface normal, and $\mathbf{p}_{\mathcal{S}}$ the nearest surface point. The first term enforces contact stability, while the second penalizes penetration below the ground plane.

The final optimization objective combines both geometric and physical consistency:
\begin{equation}
\mathcal{L}_{\text{opt}} = 
\mathcal{L}_{\text{BA}} 
+ \lambda_c \mathcal{L}_{\text{contact}},
\end{equation}
where $\lambda_c$ balances contact constraints against multi-view reprojection consistency.

%% file: sec/4_experiments.tex
\input{tab/main-experiments}
\section{Experiments}
\subsection{Dataset and Metrics}
\textbf{Datasets.}
We evaluate our framework on two large-scale multi-view datasets: EgoHumans~\cite{egohuman} and EgoExo4D~\cite{egoexo}, which provide complementary coverage of human motion and scene diversity. EgoHumans captures dynamic indoor and outdoor activities from multiple moving cameras with synchronized inertial signals and dense 3D ground truth. EgoExo4D extends this setup to sequences involving complex human–object and human–environment interactions. 

\textbf{Evaluation Metrics.}
Human metrics include:
(1) \textbf{W-MPJPE} and \textbf{WA-MPJPE}, which measure the mean per-joint position error in world coordinates after aligning the first two frames and entire sequence alignment, respectively~\cite{tram, wham};
(2) \textbf{PA-MPJPE}, the Procrustes-aligned joint error capturing shape and pose accuracy;
(3) \textbf{Accel}, the mean acceleration error reflecting temporal smoothness of the recovered motion. Camera metrics include:
(1) \textbf{TE} and \textbf{s-TE}, trajectory errors computed before and after scale normalization to evaluate global motion coherence; 
(2) \textbf{RRA@\{50,100\}} measures the fraction of view pairs whose relative pose;
(3) \textbf{CCA@\{50,100\}} quantifies the alignment accuracy between estimated camera centers and scene geometry; and
(4) \textbf{s-CCA@\{50,100\}} extends this metric after normalization to evaluate scale-invariant consistency. 

\subsection{Implementation Details}

\textbf{Scene Branch}.
Our Scene Branch is fully training-free: we keep the backbone~\cite{dust3r, monst3r, cut3r} frozen and modify only the inference procedure. We apply the special softmax function in human-region attention weights~\cite{dust3r, monst3r} to suppress dynamic interference and adopt a lightweight dual-memory update mechanism~\cite{cut3r} that disentangles same-frame multi-view fusion from cross-time background accumulation. These inference-only changes stabilize static-scene reconstruction without retraining. For human masking, we adopt the original annotations in EgoHumans~\cite{egohuman} and apply Grounded SAM 2~\cite{kirillov2023segany, ravi2024sam2segmentimages, ren2024grounded, ren2024grounding} for EgoExo4D~\cite{egoexo}.

\textbf{Human Branch}.
We train this module on 2 NVIDIA A800 GPUs using AdamW~\cite{adamw} and a batch size of 16 sequences, each with a 16-frame window.
The training consists of three stages. We adopt the weight of TRAM~\cite{tram} as the initial weight. In the first stage, we froze the initial weights and only trained the contact head for 10K steps on 3DPW~\cite{3dpw}, BEDLAM~\cite{bedlam} and Human3.6M~\cite{human3}. In the second stage, we train only the anchor-aware cross-view attention module and freeze the other weights for 20K steps on EgoHumans~\cite{egohuman}. Finally, we fine-tune all the weights on all datasets~\cite{3dpw, bedlam, human3, egohuman} for an additional 2K steps. 

\subsection{Results}
\textbf{Quantitative results.} 
We evaluate the proposed \methodName framework on the EgoHumans and EgoExo4D benchmarks, comparing it against state-of-the-art methods for human, camera and scene reconstruction.  \methodName is built upon three transformer-based dense multi-view backbones~\cite{dust3r,monst3r,cut3r} and extends them with a unified human–scene optimization and global alignment strategy.  As reported in~\cref{tab:main}, \methodName consistently improves performance across all metrics, demonstrating strong generalization across both architectures and datasets. Compared to HSfM~\cite{hsfm}, which performs human and scene reconstruction per-frame, \methodName reduces W-MPJPE by over 50\% and PA-MPJPE by more than 1.5$\times$, while also achieving the lowest acceleration error, indicating smoother and physically coherent human motion.  In addition, \methodName improves trajectory error and relative reconstruction accuracy, confirming better spatial alignment between cameras and the static environment.  The consistent gains in CCA and s-CCA further highlight improved global geometric coherence among humans, scenes, and cameras.  Importantly, all three backbones benefit consistently from our framework, validating \methodName as a backbone-agnostic and plug-and-play solution for globally consistent human–scene reconstruction. It is also worth noting that camera-related metrics on EgoExo4D are higher than those on EgoHumans.  This difference stems from the data characteristics: as of EgoHumans, we adopt the ego-view cameras mounted on moving subjects, resulting in dynamically changing viewpoints with frequent motion blur and occlusion.  In contrast, EgoExo4D provides more stable third-person camera settings.  Hence, while \methodName achieves strong results on both datasets, the EgoHumans benchmark presents a substantially more challenging setting that requires handling dynamic and uncalibrated egocentric motion.

\textbf{Qualitative Results.}
\Cref{fig:opt} visualizes three representative scenes using our DUSt3R-based integration. 
For Scenes~1 and~2, the initial reconstructions (\cref{fig:opt} a, c) exhibit misalignments between humans, scenes, and cameras, leading to physical violations such as interpenetration, floating feet, and incorrect grounding (highlighted in red boxes). 
After applying our global optimization (\cref{fig:opt} b, d), \methodName jointly refines human motion, camera poses, and scene geometry through contact-aware constraints, producing physically plausible, globally aligned reconstructions with accurate human-ground contact and consistent spatial placement across all views. Scene~3 presents a direct comparison between HSfM and our \methodName framework. 
Since HSfM performs frame-wise independent optimization, its per-frame scale estimates are not temporally harmonized, causing scale drift across time. 
When aggregated, these scale inconsistencies manifest as spatial shifts and misaligned human–scene geometry (\cref{fig:opt} e). 
In contrast, \methodName enforces a unified global scale and jointly optimizes all components across the entire sequence, yielding stable and temporally coherent reconstructions that preserve consistent human placement relative to the static environment (\cref{fig:opt} f).

Overall, these results demonstrate that \methodName advances multi-view human–scene understanding by integrating cross-view feature interaction, human-aware attention and physically grounded global optimization. 
The method delivers both numerical gains and significantly improved visual fidelity, and generalizes robustly across different architectures and dataset domains without additional training.

\input{tab/multi-view}

\subsection{Ablation Study}
\textbf{Human Branch}.
To evaluate the effectiveness of the proposed Human Branch, we conduct ablation studies on the Multi-View dataset, as shown in~\cref{tab:multi-view}. We compare our method with several state-of-the-art video-based pose estimators, including HMR2.0~\cite{hmr, hmr2}, GVHMR~\cite{gvhmr}, and VIMO~\cite{tram}. All baselines are trained on monocular inputs and do not explicitly enforce multi-view geometric consistency. 

Even without gravity-aware alignment, our model outperforms all baselines across all metrics, reducing MPJPE from $81.2$ (VIMO ft) to $79.1$ and improving acceleration smoothness from $17.9$ to $17.6$. When incorporating the gravity constraint, performance further improves to $77.7$ MPJPE and $16.8$ accel, achieving the best accuracy and temporal stability among all methods. These results demonstrate that our symmetric and anchor-referenced cross-view attention effectively fuses geometric cues across synchronized views, leading to more consistent pose estimation. The gravity-aware term further enhances physical plausibility by stabilizing vertical motion and suppressing drift. Overall, our Human Branch enables globally consistent, physically grounded human motion reconstruction that complements the static scene geometry estimated by the Scene Branch.
\input{tab/scene}
\textbf{Scene Branch}.
We evaluate the effectiveness of our plug-and-play human-aware attention module across three representative transformer-based reconstruction backbones: DUSt3R~\cite{dust3r}, MonST3R~\cite{monst3r}, and CUT3R~\cite{cut3r}. As shown in~\cref{tab:scene}, integrating our module consistently improves geometric stability and reconstruction accuracy under dynamic scenes. For DUSt3R and MonST3R, we adopt a masked attention strategy that suppresses the influence of dynamic human regions during feature aggregation. For CUT3R, the same concept is realized through a multi-memory disentanglement mechanism that separates static and dynamic information across time and views. Across all backbones, introducing human awareness yields a noticeable reduction in Trajectory Error (TE) and Alignment Error (AE), with average improvements of 4–6\% over the baseline models. The RRA@100 and s-CCA@100 metrics also increase consistently, demonstrating enhanced spatial alignment and temporal coherence of reconstructed geometry. Notably, CUT3R benefits the most from our module (TE from 1.90 to 1.83; s-CCA from 0.48 to 0.52), reflecting its strong synergy with our multi-memory design. These results confirm that the proposed module serves as a universal enhancement for transformer-based multi-view reconstruction frameworks. By implicitly suppressing human-induced inconsistencies without requiring training, it enables more stable, physically consistent, and human-aware scene reconstructions across diverse backbone architectures.

%% file: tab/main-experiments.tex
\begin{table*}[htbp]
\centering
\caption{\textbf{Evaluation of \methodName on the EgoHumans~\cite{egohuman} and EgoExo4D~\cite{egoexo} datasets.} 
Across all backbones and datasets, \methodName consistently improves human accuracy (lower W-MPJPE, PA-MPJPE, Accel) and enhances scene–camera coherence (higher RRA, CCA, s-CCA), outperforming the original baselines by a large margin. These results demonstrate the generalization of our framework across architectures and domains, yielding globally aligned and physically coherent human–scene reconstructions.}
\vspace{-6pt}
\resizebox*{1.0\textwidth}{!}{
\begin{tabular}{clcccccccccccc}
\hline
\multicolumn{2}{c}{\multirow{2}{*}{Method}}& \multicolumn{4}{c}{\textbf{Human Metrics}} & \multicolumn{8}{c}{\textbf{Camera Metrics}}  \\
\multicolumn{2}{c}{} & W-MPJPE$\downarrow$ & WA-MPJPE$\downarrow$  & PA-MPJPE$\downarrow$  & Accel$\downarrow$ & TE$\downarrow$   & s-TE$\downarrow$  & RRA@50$\uparrow$  & RRA@100$\uparrow$  & CCA@50$\uparrow$  & CCA@100$\uparrow$  & s-CCA@50$\uparrow$  & s-CCA@100$\uparrow$  \\ \hline
\multicolumn{1}{c|}{\multirow{7}{*}{\rotatebox{90}{EgoHumans}}} & DUSt3R~\cite{dust3r} & - & - & - & - & 3.20 & 1.37 & 0.45 & 0.65 & - & - & 0.22 & 0.49 \\
\multicolumn{1}{c|}{} & MonST3R~\cite{monst3r} & - & - & - & - & 3.17 & 1.47 & 0.44 & 0.66 & - & -  &  0.20   &    0.46      \\
\multicolumn{1}{c|}{} & CUT3R~\cite{cut3r} & - & - & - & - & 1.90 & 1.44 & 0.44 & 0.65 & 0.06 & 0.38  &     0.20   &    0.48      \\
\multicolumn{1}{c|}{} & HSfM*~\cite{hsfm}  & 227.82 & 163.97  &  21.93  &  57.89 &  1.79 & 1.23  & 0.49 & 0.70 &  0.12  &  0.42    &  0.26   &  0.52   \\ 
\multicolumn{1}{c|}{} & \textbf{\methodName (DUSt3R)}   &  106.31 &   \textbf{69.22} & 22.74   &  \textbf{13.74}    &    1.31   &  1.07   &  \textbf{ 0.57}  &  \textbf{0.79}    &   0.24     &     0.58   &   0.38  &  \textbf{0.63}        \\
\multicolumn{1}{c|}{} & \textbf{\methodName (MonST3R)}  &  105.17  &  78.74  &  26.95   &   14.81    &   1.38    &  1.02  &   0.54     &  0.76  &    0.22    &    0.57  &     0.39      &     0.61     \\
\multicolumn{1}{c|}{}  & \textbf{\methodName (CUT3R)}   &  \textbf{97.54}  & 73.23  &  \textbf{20.71} & 14.23 & \textbf{1.03} & \textbf{0.86} & 0.55 &   0.78   &   \textbf{0.26} &    \textbf{0.59}     &  \textbf{0.40}  &   \textbf{0.63}   \\
 \hline
\multicolumn{1}{c|}{\multirow{7}{*}{\rotatebox{90}{EgoExo4D}}} & DUSt3R~\cite{dust3r} & - & - & - & - & 4.84 & 1.22 &0.86 & 0.92 & - & - & 0.13 & 0.58\\
\multicolumn{1}{c|}{} & MonST3R~\cite{monst3r} & - & - & - & - & 4.83 & 0.55 & 0.91 & 0.98 & - & -  & 0.78  &  0.90     \\
\multicolumn{1}{c|}{} & CUT3R~\cite{cut3r} & - & - & - & - & 1.93 & 0.21 & 0.92 & 0.95 & 0.05 & 0.12 & 0.91 & 0.96     \\
\multicolumn{1}{c|}{} & HSfM*~\cite{hsfm}  & 123.12 & 109.23  & 17.82  & 49.27  & 2.85 & 0.96  & 0.90 & 0.94 & 0.19  & 0.25 & 0.84 & 0.91 \\
\multicolumn{1}{c|}{} & \textbf{\methodName (DUSt3R)}   & 112.78 & 72.41   &  15.31  & 16.12  &  2.37  & 0.92  &  0.90   &   0.96   &    0.17    &   0.25   &   0.67  &   0.73      \\
\multicolumn{1}{c|}{} & \textbf{\methodName (MonST3R)}  & 97.13  & 76.16  & \textbf{14.23} & \textbf{15.17}   &  2.61  & 0.55 & \textbf{0.97}  & \textbf{1.00}   &  0.19    &  0.26      &     0.81   &   0.93       \\
\multicolumn{1}{c|}{}  & \textbf{\methodName (CUT3R)}   & \textbf{91.7} & \textbf{70.29} &  16.92 &16.72 & \textbf{1.38} & \textbf{0.18} & \textbf{0.97} & 0.98 &   \textbf{0.24}     &  \textbf{0.29}     &   \textbf{0.97}     &    \textbf{0.99}    \\
 \hline
\end{tabular}
}
\vspace{-10pt}
\label{tab:main}
\end{table*}

%% file: tab/multi-view.tex
\begin{table}[t]
\setlength{\tabcolsep}{3pt}
\centering
\caption{\textbf{Evaluation results of human branch on the EgoHuman dataset.} 
We compare it with recent human reconstruction approaches under both the \emph{All Views} and \emph{Anchor View}. 
Across all metrics, \methodName (human branch) outperforms prior methods.
}
\resizebox{0.47\textwidth}{!}{
\begin{tabular}{clccll}
\hline
\multicolumn{2}{c}{Method}   & PA-MPJPE$\downarrow$   & MPJPE$\downarrow$   & PVE$\downarrow$ & Accel$\downarrow$  \\ \hline
\multicolumn{1}{c|}{\multirow{6}{*}{\rotatebox{90}{All Views}}}   & HMR2.0~\cite{hmr,hmr2}  & 57.6 & 107.6 & 121.2 &  46.5                       \\
\multicolumn{1}{c|}{}   & GVHMR~\cite{gvhmr}     &  58.7 & 108.5 & 125.6  &  44.7          \\
\multicolumn{1}{c|}{}  & VIMO~\cite{tram} (w/o finetuning) & 46.3  &  89.8   & 112.3  &   19.8              \\
\multicolumn{1}{c|}{}  & VIMO~\cite{tram} (with finetuning) & 41.4 &  81.2   & 108.2  &  17.9    \\ \cline{2-6} 
\multicolumn{1}{c|}{}  & \textbf{\methodName (human branch, w/o gravity)}  &   39.2        &  79.1         &   99.7          &    17.6   \\
\multicolumn{1}{c|}{}  & \textbf{\methodName (human branch, with gravity)}   &    \textbf{38.8}        &  \textbf{77.7}         &   \textbf{98.1}          &    \textbf{16.8}         \\ \hline
\multicolumn{1}{c|}{\multirow{6}{*}{\rotatebox{90}{Anchor View}}} & HMR2.0~\cite{hmr,hmr2}  &  58.2   &  109.4   &  125.3 &  47.3 \\
\multicolumn{1}{l|}{}  & GVHMR~\cite{gvhmr}   &  57.9          &  109.1   & 124.3  &  45.2     \\
\multicolumn{1}{l|}{}   & VIMO~\cite{tram} (w/o finetuning)   &  47.7   &  91.4   & 116.3  &   20.1 \\
\multicolumn{1}{l|}{}    & VIMO\ddag~\cite{tram} (with finetuning)   &  42.3   &  81.9   & 109.7  & 18.3   \\ \cline{2-6} 
\multicolumn{1}{l|}{}   & \textbf{\methodName (human branch, w/o gravity)}   &     37.9           &   77.3  & 97.4  &   17.3  \\
\multicolumn{1}{l|}{}    & \textbf{\methodName (human branch, with gravity)}     &     \textbf{36.2}           &   \textbf{75.7}  & \textbf{94.9}  &   \textbf{13.6}  \\ \hline
\end{tabular}}\label{tab:multi-view}
\vspace{-10pt}
\end{table}

%% file: tab/scene.tex
\begin{table}[t]
\small
\caption{\textbf{Ablation Study on the human-aware attention.} Across all backbones~\cite{cut3r, monst3r, dust3r}, introducing human-aware attention to the scene branch consistently improves TE, AE and RRA, indicating more stable and geometrically consistent reconstruction.}
\resizebox{0.47\textwidth}{!}{
\begin{tabular}{lccccccc}
\hline
& TE$\downarrow$ & s-TE$\downarrow$ & AE$\downarrow$   & RRA@100$\uparrow$ & s-CCA@100$\uparrow$ \\ \hline
DUSt3R~\cite{dust3r}   & 3.20 & 1.37 & 107.21 & 0.65 & 0.49      \\
+ human-aware attention  & \textbf{3.05}  &  \textbf{1.32}  & \textbf{104.93} & \textbf{0.69} & \textbf{0.52}   \\ \hline

MonST3R~\cite{monst3r}  & 3.17 & 1.47 & 108.82 & 0.66 &   0.46    \\
+ human-aware attention  & \textbf{3.06}   &   \textbf{1.36}   & \textbf{101.71} &  \textbf{0.67} &  \textbf{0.51}    \\ \hline

CUT3R~\cite{cut3r}  & 1.90 & 1.44 & 106.92 & 0.65 & 0.48   \\
+ human-aware attention  & \textbf{1.83} & \textbf{1.33}  & \textbf{103.34} & \textbf{0.68}  &   \textbf{0.52}   \\ \hline
\end{tabular}} \label{tab:scene}
\vspace{-10pt}
\end{table}

%% file: sec/5_con.tex
\section{Conclusion}
We present \methodName, a unified framework for consistent human-scene reconstruction from multi-view inputs. 
By jointly leveraging dense geometric reasoning and articulated human modeling,
\methodName bridges scene and motion estimation through a shared optimization backbone.

%% file: sec/6_ack.tex
\section*{Acknowledgment}

This work is partially supported by the Ministry of Education, Singapore, under the Academic Research Fund Tier 1 (FY 2025), National University of Singapore Robotics Seed Grant, and by a research gift from Futurewei Technologies Inc.